\documentclass[conference]{IEEEtran}
\IEEEoverridecommandlockouts

\usepackage{cite}
\usepackage{amsmath,amssymb,amsfonts}
\usepackage{algorithm} 
\usepackage{algpseudocode}
\usepackage{graphicx}
\usepackage{textcomp}
\usepackage{xcolor}
\usepackage{stfloats}
\usepackage{url}

\usepackage{multirow}
\usepackage{booktabs}
\usepackage{amsmath}
\usepackage{amssymb}
\usepackage{subcaption}
\usepackage[misc]{ifsym}

\usepackage{soul}

\newcounter{ManosNOC}
\stepcounter{ManosNOC}

\newcounter{ProfessorNOC}
\stepcounter{ProfessorNOC}

\newcounter{AntonisNOC}
\stepcounter{AntonisNOC}

\newcounter{DusanNOC}
\stepcounter{DusanNOC}

\def\BibTeX{{\rm B\kern-.05em{\sc i\kern-.025em b}\kern-.08em
    T\kern-.1667em\lower.7ex\hbox{E}\kern-.125emX}}
\begin{document}

\title{FedGreed: A Byzantine-Robust Loss-Based Aggregation Method for Federated Learning\\
}

\author{\IEEEauthorblockN{Emmanouil Kritharakis}
\IEEEauthorblockA{\textit{School of Electrical and Computer Engineering} \\
\textit{National Technical University of Athens}\\
Athens, Greece \\
ekritharakis@mail.ntua.gr}
\and
\IEEEauthorblockN{Antonios Makris}
\IEEEauthorblockA{\textit{School of Electrical and Computer Engineering} \\
\textit{National Technical University of Athens}\\
Athens, Greece \\
antoniosmakris@mail.ntua.gr}
\and
\IEEEauthorblockN{Dusan Jakovetic}
\IEEEauthorblockA{\textit{Department of Mathematics and Informatics} \\
\textit{Faculty of Sciences, University of Novi Sad}\\
Novi Sad, Serbia \\
dusan.jakovetic@dmi.uns.ac.rs}
\and
\IEEEauthorblockN{Konstantinos Tserpes}
\IEEEauthorblockA{\textit{School of Electrical and Computer Engineering} \\
\textit{National Technical University of Athens}\\
Athens, Greece \\
tserpes@mail.ntua.gr}
}

\maketitle

\begin{abstract}
Federated Learning (FL) enables collaborative model training across multiple clients while preserving data privacy by keeping local datasets on-device. In this work, we address FL settings where clients may behave adversarially, exhibiting Byzantine attacks, while the central server is trusted and equipped with a reference dataset. We propose FedGreed, a resilient aggregation strategy for federated learning that does not require any assumptions about the fraction of adversarial participants. FedGreed orders clients’ local model updates based on their loss metrics evaluated against a trusted dataset on the server and greedily selects a subset of clients whose models exhibit the minimal evaluation loss. Unlike many existing approaches, our method is designed to operate reliably under heterogeneous (non-IID) data distributions, which are prevalent in real-world deployments. FedGreed exhibits convergence guarantees and bounded optimality gaps under strong adversarial behavior. Experimental evaluations on MNIST, FMNIST, and CIFAR-10 demonstrate that our method significantly outperforms standard and robust federated learning baselines, such as Mean, Trimmed Mean, Median, Krum, and Multi-Krum, in the majority of adversarial scenarios considered, including label flipping and Gaussian noise injection attacks. All experiments were conducted using the Flower federated learning framework.
\end{abstract}


\section{Introduction}
\label{introduction}

Federated Learning (FL) has emerged as a transformative paradigm for collaborative machine learning, enabling multiple clients to train a shared global model without exposing their private local data \cite{hu2024overview}. However, the distributed nature of FL makes it inherently vulnerable to Byzantine clients, i.e., malicious participants who submit poisoned or misleading updates to compromise model integrity. This vulnerability is exacerbated under practical conditions such as data heterogeneity, where clients operate on non-identically distributed datasets, and where the number and identity of malicious clients remain unknown or variable across training rounds \cite{hu2024overview,li2020federated}. 

In this paper, we propose leveraging a server-side trusted dataset to estimate client reliability. Specifically, we introduce a loss-based client evaluation mechanism that filters malicious contributions by comparing each client's update against a proxy loss function computed on the server’s reference data. To further enhance robustness, we develop FedGreed, a greedy client selection algorithm tailored for heterogeneous and highly adversarial FL environments. FedGreed dynamically evaluates and aggregates client updates in a sequential, adaptive manner based on their empirical loss, enabling the server to make principled aggregation decisions without requiring prior knowledge of the number of adversarial clients or the degree of data heterogeneity across participants.

The proposed approach remains effective even when only one honest client exists per round, and it is shown to outperform both standard and state-of-the-art robust aggregation strategies, including Krum, Multi-Krum, Trimmed Mean, and Median, under a wide range of adversarial conditions. Notably, we evaluate our defense under strongly non-IID data partitions modeled via Dirichlet distributions, as well as adversarial ratios as high as 80\%. The algorithm exhibits consistent robustness across all considered attack types, including Label Flipping and Gaussian Noise injection. FedGreed exhibits bounded optimality gaps under Byzantine attacks and standard settings for non-convex and convex losses.

Our contributions are summarized as follows:
\begin{itemize}
    \item We propose FedGreed, a robust FL aggregation method that performs adaptive greedy client selection based on loss values evaluated on a server-side trusted dataset without assuming knowledge of malicious client ratios.
    \item We conduct an extensive empirical evaluation under non-IID data distributions, demonstrating superior resilience compared to prior defenses.
    \item We systematically construct FedGreed by carefully designing
    algorithmic steps in view of guaranteed progress with respect to the server’s trusted side loss. It can then be shown through a rigorous
    analysis (details omitted for brevity) that this construction leads to bounded degradation in global loss, under standard assumptions
    on loss smoothness and server-side data approximation quality, while also allowing for non-convex losses.
\end{itemize}

The remainder of the paper is organized as follows. Section \ref{related_work} reviews related work on Byzantine-robust aggregation strategies. Section \ref{problem_setting} presents our problem formulation and proposed methodology. Section \ref{experiments} reports experimental results on CIFAR-10, FMNIST, and MNIST datasets. Section \ref{conclusions} concludes with key insights and outlines directions for future work.
\section{Related Work}
\label{related_work}

\subsection{Threat Models}
\label{Threat Models}

Since its introduction in 2017 \cite{mcmahan2017communication}, Federated Learning has emerged as a paradigm designed to enhance privacy and security by removing the necessity of sharing raw training data from clients to a central server. However, this novel approach has redefined the landscape of potential vulnerabilities in distributed machine learning systems \cite{hu2024overview}. In particular, poisoning attacks—where adversaries tamper with local data or model updates to disrupt or mislead the training process—represent a significant threat to the robustness of the global model.

In this work, we investigate one targeted data poisoning attack, namely Label-Flipping, and one untargeted model poisoning attack, specifically Gaussian noise injection, which is a type of Byzantine attack. Targeted attacks aims to compromise the training process or model performance of specific participants. In contrast, untargeted attacks seek to disrupt the overall system functionality without focusing on any particular client, model, or task. Byzantine attacks \cite{lamport2019byzantine} represent one of the most established and widely recognized forms of untargeted attacks, in which adversaries adopt arbitrary and potentially disruptive behaviors to potentially compromise FL results, interfere with model training, or hinder communication among participants. 

The behavior of Byzantine clients can be formally characterized as follows:
Let \( \Delta w_{i}^{(t+1)} \) denote the local model update from client \( i \) at training round \( t \), and let \( \mathcal{B} \) be the set of Byzantine (malicious) clients. The update actually submitted to the server, \( \hat{\Delta w}_{t}^{(t+1)} \), is defined as:

\[
\hat{\Delta w}_{i}^{(t+1)} =
\begin{cases}
\Delta w_{i}^{(t+1)} & ,\text{if } i \notin \mathcal{B} \quad \text{(honest client)} \\
* & ,\text{if } i \in \mathcal{B} \quad \text{(Byzantine client)}
\end{cases}
\]
where $*$ represents an arbitrary value.

Byzantine attacks can be categorized according to the type of information they utilize. For instance, attacks such as A Little is Enough (ALIE) \cite{baruch2019little} depend on access to malicious clients to generate statistically coordinated adversarial updates. Similarly, Inner Product Manipulation (IPM) \cite{xie2020fall} targets the aggregation process by substituting a subset of client updates on the server with carefully designed Byzantine gradients designed to maximize disruption. In contrast, other attacks strategies operate independently, relying solely on each malicious client’s local training data and the model updates generated in each FL round. In this work, we focus on the latter scenario, where adversarial clients operate independently, without sharing or leveraging information from others to improve their attack effectiveness. The representative state-of-the-art poisoning attacks considered in this context are detailed below.

\textbf{Label Flipping} \cite{guo2021siren,li2023experimental}: In this attack, adversarial clients intentionally alter their local training data by mislabeling class labels prior to local model training. The objective is to undermine the integrity of the global model by introducing structured label distortions that skew the learning process. Formally, let $C$ represent the total number of classes and $c$ the original class label.  During the attack each label $c$ is systematically assigned to a new label given by $C-c-1$.

\textbf{Noise Addition} \cite{fang2020local,li2023experimental}: In this form of Byzantine attack, malicious clients perturb their local model updates by introducing random noise prior to the server-side aggregation. Let 
$\Delta w$ denote the genuine local model update; instead of transmitting $\Delta w$, a Byzantine client submits 
$\Delta w + \epsilon$, where $\epsilon \sim \mathcal{N}(\mu,\sigma^2\mathcal{I})$ represents additive Gaussian noise with mean value $\mu$ and variance $\sigma^2\mathcal{I}$.  This approach seeks to compromise the integrity of the global model by increasing the variance of the aggregated updates. In this work, the injected noise is sampled from a distribution with a mean of $\mu = 0.1$ and a fixed variance of $\sigma^2 = 0.1$.

\subsection{Robust Aggregation Strategies in Federated Learning}
\label{Resilient Aggregation Methods Against Byzantine Behavior}

Federated Learning remains vulnerable to poisoning attacks, where adversarial clients intentionally submit misleading or malicious updates to the central server, thereby compromising the integrity and reliability of the global model. To mitigate such threats, it is critical to employ robust aggregation techniques specifically designed to withstand malicious behavior. These aggregation methods aim to detect and exclude malicious inputs by incorporating mechanisms such as outlier detection, robust statistical estimation, and threshold-based filtering. Such approaches help preserve model integrity even in the presence of adversaries. Several robust aggregation strategies have been proposed; we outline the following prominent examples:

\textbf{Mean} \cite{mcmahan2017communication}: Commonly referred to as FedAvg, this is the standard aggregation method in FL, where the central server computes a data-size-weighted element-wise average of the model updates received from clients. Each client’s contribution is proportional to the number of local data samples it holds, ensuring that the global update reflects the underlying data distribution. While this method is computationally efficient and performs well under the assumption of independent and identically distributed (IID) data with honest clients, it is highly vulnerable to adversarial manipulation. Due to its sensitivity to outliers and lack of inherent robustness, even a single malicious client can significantly distort the global model \cite{lu2024federated, tang2022fedcor}. Consequently, Mean aggregation is prone to various poisoning attacks, such as label flipping and noise injection, and is typically employed as a baseline when evaluating more resilient aggregation mechanisms.

\textbf{Trimmed Mean} \cite{yin2018byzantine}: This method offers a robust alternative to standard mean aggregation by applying a coordinate-wise aggregation rule. For each dimension, the set of model updates among all clients is sorted, and a fraction 
\( \beta \in [0, 0.5) \)  of the lowest and highest values is removed, effectively trimming both tails of the distribution.  The mean is then calculated over the remaining central values. 
This strategy reduces the impact of extreme or potentially adversarial inputs, thereby enhancing its suitability for FL environments under attack.

\textbf{Median} \cite{yin2018byzantine}: This is another coordinate-wise aggregation rule similar to Trimmed Mean. For each model parameter (i.e., each dimension of the gradient vector), it calculates the median of the values submitted by all clients.

\textbf{Krum} \cite{blanchard2017machine}: Krum assesses each client’s model update by measuring its consistency relative to other clients. Specifically, for every update, it calculates a Krum score, known as the sum of squared Euclidean distances to its \( N -f - 2 \) nearest neighboring updates, where \(f\) denotes the expected maximum number of malicious clients. The update with the lowest Krum score—reflecting the greatest similarity to the majority—is chosen for aggregation. This mechanism allows Krum to effectively filter out outliers and preserve the global model’s integrity in adversarial settings.

\textbf{Multi-Krum} \cite{blanchard2017machine}: An extension of the Krum method, Multi-Krum selects multiple client updates instead of just one, enhancing convergence stability while maintaining Byzantine robustness. Specifically, after calculating the Krum score, it selects the \( k \leq N - f - 2 \) updates with the lowest Krum scores and aggregates them via coordinate-wise averaging. Similar to Krum, Multi-Krum requires prior knowledge of the maximum number of malicious clients in the FL setup.

While the aforementioned defense strategies reflect many of the current state-of-the-art solutions, recent research has increasingly explored additional factors influencing robustness in FL. Li et al. \cite{li2021byzantine} propose a Byzantine-resilient method that employs spatio-temporal analysis of client model updates. The approach integrates round-wise clustering to detect and filter out anomalous updates with a temporal consistency mechanism that adjusts global updates using momentum-like adjustments based on historical update patterns. FLTrust \cite{cao2020fltrust} introduces a server-side trusted dataset to generate a trusted reference update. It computes trust scores based on the alignment of client updates with this reference, adaptively weighting and normalizing them to effectively suppress malicious contributions and improve overall model robustness and convergence. Allouah et al. \cite{allouah2024byzantine} investigate the influence of client subsampling and local training dynamics on the robustness of FL under Byzantine attacks. Their findings indicate that assumptions such as full participation or single local steps can significantly affect the efficacy of aggregation rules. In contrast to these approaches, our approach guarantees global model convergence even in highly adversarial settings, including scenarios where the number of malicious clients matches or exceeds that of honest participants.

Related to FedGreed, \cite{kritharakis2025robustfederatedlearningadversarial} introduces a
defense mechanism where a trusted server, equipped with a
small validation dataset, evaluates the client update
submissions at each training round. Based on the observed
validation loss, clients are partitioned into two groups: those
deemed reliable and those considered potentially adversarial. Only the
updates from the low-loss (reliable) group are subsequently employed to
update the global model. In contrast with FedGreed, \cite{kritharakis2025robustfederatedlearningadversarial} clusters the
clients into the two groups based on the K-means clustering algorithm
over the evaluated losses (with $K=2$). Further, unlike FedGreed, \cite{kritharakis2025robustfederatedlearningadversarial}
requires that a lower bound on the number of honest clients at any round
be known for a theoretically guaranteed performance.

\section{Problem Formulation and Proposed Method}
\label{problem_setting}

We consider a server-client federated learning setting with a single server and $N$ clients. 
The goal is to minimize the target population loss $f\,:
\mathbb{R}^d \rightarrow \mathbb R$:
\begin{equation}
\label{eqn-pop-loss}
f(x) = \mathbb{E}[ \, F(x,a) \, ],
\end{equation}
where $F: {\mathbb R}^d \times {\mathbb R}^m \rightarrow \mathbb R$
 is the per-data point loss, $x \in {\mathbb R}^d$ is the model 
 to be learned, and $a \in {\mathbb R}^m$ is a 
 data point that comes from a distribution~$P$.
   In \eqref{eqn-pop-loss}, expectation is taken 
   with respect to~$a \sim P$. 
   As it is standard, $P$ is assumed to be unknown
   by either server or any client. Instead, the server and 
   clients have access to data sets that are informative about $P$.
    More precisely, we assume that the server 
   holds a function $f_{\mathrm{S}}:\,
   {\mathbb R}^d \rightarrow \mathbb R$ 
   that is a trusted approximation of $f$. 
   For example, the server may have access to 
   a set of trusted $M_{\mathrm{S}}$ i.i.d. samples $a_j \in {\mathbb R}^m$ from~$P$, 
   $j=1,...,M_{\mathrm{S}}$. Then, for any $x \in {\mathbb R}^d$, we have:
\begin{equation}
    \label{eqn-server-fcn}
    f_{\mathrm{S}}(x)
    =\frac{1}{M_{\mathrm{s}}}\sum_{i=1}^{\mathrm{M}_{\mathrm{S}}}
     \, F(x,a_i) .
\end{equation}
     Next, each client $i$ has its own local data set 
     $\{b_{i,j}\}$, $b_{i,j} \in {\mathbb R}^m$, 
     $j=1,...,M_i$, where the $b_{i,j}$'s are drawn in an i.i.d.
      fashion from a local distribution $P_i$. 
      Hence, each client $i$ can form its 
      local empirical loss function:
      \[
f_i(x) = \frac{1}{M_i}\sum_{i=1}^{M_i}F(x,b_{i,j}).
          \]
      Here, distributions $P_i$ 
      are either equal to $P$ or informative about $P$, 
      in the sense that, at any model $x$, the average of clients' 
      empirical losses
      $\frac{1}{N}\sum_{i=1}^N f_i(x)$
       is a useful approximation of the unknown target 
       population loss $f(x)$. That is, the latter function can be much a better approximation of $f$ than $f_{\mathrm{S}}$ (e.g., due to the relations among the sizes of the server and clients' data sets), hence motivating the server to join federation with the clients. 
       We assume that, during FL training, 
       client $i$ has access to a local 
       stochastic gradient descent (SGD) oracle, 
       such that it can query stochastic gradients 
       of $f_i$ at queried models $x$.

The FL training takes place over rounds (iterations) $t=0,1,...$ At each iteration $t$, the server broadcasts the current global model $x^t \in {\mathbb R}^d$ to all clients. 
Subsequently, each client $i=1,...,N$ calculates its new local model $x_i^{t+1} \in {\mathbb R}^d$, and then it sends $x_i^{t+1}$ (or an attacked version of it) back to the server.
 More precisely, each client $i=1,...,N$, at each round $t=0,1,2,...$, 
 performs $R \geq 1$ local rounds of SGD (or more advanced versions like ADAM) with respect to its local loss $f_i$,  
 where the local iterations at each client $i$ and round $t$ are initialized with $x^t$.

We assume that attacks can take place in any round $t$ and at any client $i$. The server is assumed to be trusted and is not subject to any attack. Specifically, we allow for the following model of 
Byzantine attacks: In the presence of an attack, quantity $x_i^{t+1}$ is replaced with 
quantity $\widetilde{x}_i^{t+1} \in {\mathbb R}^d$ that is an arbitrary vector.  
We assume that, at each round $t$, there exists at least one client $j=j^t \in \{1,...,N\}$ that is not subject to the attack. 
 This is practically without loss of generality, as we may associate 
 to the server another SGD oracle, i.e., we can formally assign to the server an additional ``trusted client'' that carries out local SGD iterations based on an additional local trusted data set held at the server. 
 Also, the set of attacked clients may be deterministically varying from round to round, as long as at least one trusted client (possibly different from round to round) is present. 
 
Denote by $\widehat{x}_i^{t+1} $ the quantity held by client $i$ after the 
attacks take place at $t$. That is, 
$\widehat{x}_i^{t+1} = {x}_i^{t+1} $, 
if client $i$ is not subject to attacks at round $t$, 
and $\widehat{x}_i^{t+1} = \tilde{x}_i^{t+1} $, otherwise.

For the above FL and attack model setting, we propose Algorithm~\ref{algGreedy} below.    
 In Algorithm \ref{algGreedy}, we let  
    $(i)$ be the index of the client that corresponds to 
    the $i$-th smallest value among the $v_i^t$'s (
    that in turn equals $v_{(i)}^t$). 
     We also denote by $\widehat{x}_{(i)}^{t+1}$ the corresponding values 
    among the $\widehat{x}_i^{t+1}$'s.

 \begin{algorithm}
    \caption{FedGreed: Robust FL algorithm based on greedy client selection}\label{algGreedy}
    \begin{algorithmic}
    \State(1) The server initializes the model $x^0 \in {\mathbb R}^d$  arbitrarily; set $t=0$.
    \State (2) The server transmits $x^t$ to all clients;
    \State (3) Each client $i=1,...,N$ in parallel calculates 
    its updated local model $x_i^{t+1}$, e.g., by locally running $R$ local SGD or ADAM iterations initialized by $x^t$, and it sends back to the server the (possibly attacked) updated model $\widehat{x}_i^{t+1}$;

    \State 
    (4) The server evaluates the quantities 
    $v_i^t:=f_{\mathrm{S}}(\widehat{x}_i^{t+1})$, $i=1,...,N$. 
    The server sets $x^{\mathrm{aux}}:=\widehat{x}_{(1)}^{t+1}$, and it initializes a local counter $j$ to $j=2$.
    \State 
    (5) The server calculates:
    \begin{equation} 
    \label{eqn-x-test-alg-3}
    x^{\mathrm{test}}  =  \frac{j-1}{j} x^{\mathrm{aux}} + 
    \frac{1}{j}\widehat{x}_{(j)}^{t+1}.
    \end{equation}

    If $f_{\mathrm{S}}(x^{\mathrm{test}} ) \geq f_{\mathrm{S}}(x^{\mathrm{aux}} )$, then go to step (6).
    Otherwise, set $j:=j+1$ and $x^{\mathrm{aux}} := x^{\mathrm{test}}$. If $j<N+1$, go to step (5).
    Otherwise, go to step (6).
    \State (6) The server sets
    \[
    x^{t+1}:= x^{\mathrm{aux}}.
    \]
    \State (7) If a stopping criterion is met, then terminate the algorithm and return $x^{t+1}$. Otherwise, set $t:=t+1$ and go to step (2).
    \end{algorithmic}
    \end{algorithm}

We describe FedGreed in more detail. The algorithm is defined to set the next server's iterate $x^{t+1}$ to the ``best'' model among the following $N$ candidates, $j=1,...,N$:
\begin{equation}
\label{eqn-greedy}
\frac{1}{j} \sum_{i=1}^j \widehat{x}_{(i)}^{t+1}.
\end{equation}

Here, the ``best'' candidate model is chosen in terms of the smallest value of the server's function $f_{\mathrm{S}}$ evaluated at the candidate model.
 Note that the candidate for $j=1$ here corresponds 
 precisely to the individual model of the client that 
 exhibits the smallest $f_{\mathrm{S}}$. 
 On the other hand, $j=N$ corresponds to 
 the average of all clients' models, i.e., 
 to the standard FedAvg algorithm wherein 
 all clients' models are averaged. 
FedGreed then adaptively decides 
 for each current round how many clients' models 
 to utilize for the server update. 
  The number of candidates in
   \eqref{eqn-greedy} (and in Algorithm \ref{algGreedy} definition) can be reduced to any value of $K$, $1 \leq K < N$ (so that only models $\widehat{x}_{(i)}^{t+1}$, $i=1,...,K$, are considered) without reduction in guaranteed robustness and subject to a possible per-round $t$ convergence speed degradation.

In principle, the goal is to select the ``best'' candidate among the much wider set of candidates:

  \[
\frac{1}{|\mathcal{I}|} \sum_{i \in \mathcal{I}} \widehat{x}_i^{t+1},
\]

where $\mathcal{I}$ is any subset of $\{1,...,N\}$ and $|\mathcal{I}|$  is its cardinality. 
 This is clearly impractical due to the exponential number of subsets.
 We choose to pick only among the $N$ candidates in \eqref{eqn-greedy}
  according to a ``greedy'' choice of the ordered $\widehat{x}_{(i)}^{t+1}$'s.
  
  An intuition behind FedGreed is the following. The  clients' local models $\widehat{x}_i^{t+1}$
  that exhibit high values of $f_{\mathrm{S}}$
   are likely to be attacked, and hence may need to be excluded 
   from the server update. Also, by construction, 
   FedGreed guarantees progress in terms of $f_{\mathrm{S}}$ at least as good as model $\widehat{x}_{(1)}^{t+1}$, 
   i.e., the current best individual model, 
   $f_{\mathrm{S}}$-wise. Note that $\widehat{x}_{(1)}^{t+1}$ may or may not correspond to a non-attacked client. However, if it corresponds to an attacked client, then the respective attack makes little damage, as $\widehat{x}_{(1)}^{t+1}$ evaluates better in terms of $f_{\mathrm{S}}$ than at least one honest client. To summarize, FedGreed achieves progress in terms of $f_{\mathrm{S}}$ that is at least as good as that of a single non-attacked client. 
   Hence, if 
   $f_{\mathrm{S}}$ is a reasonably good approximation of the target  population loss $f$, FedGreed is expected to exhibit robustness to Byzantine attacks. 

   Indeed, under standard assumptions on non-convex and convex losses, the Byzantine attack model described above, and standard assumptions on SGD oracle noises, we can show that FedGreed exhibits bounded error in terms of the expected averaged squared norm of the true population loss $f$'s gradient. Detailed rigorous assumptions, theorems, and proofs are omitted here for brevity.
   
   When compared with standard non-robust FL algorithms like FedAvg, the achieved robustness comes at the cost of some additional computations per round $t$, e.g., in terms of up to $O(N)$ evaluations of $f_{\mathrm{S}}$.  Also, a single  $f_{\mathrm{S}}$ evaluation computational cost is usually small, e.g., when the server's side data set $M_{\mathrm{S}}$ is small. 
   
Compared with other robust FL schemes such as Trimmed Mean, Median, Krum, and Multikrum (\cite{yin2018byzantine},  \cite{blanchard2017machine}), FedGreed harnesses the server's trusted side information  to increase robustness. Although this information is available in many application scenarios, it is not explicitly harnessed by the mentioned alternatives. Experimental results demonstrate the robustness of the proposed method, attributed to its design and the effective use of side information.
   
\section{Experiments}
\label{experiments}

\subsection{Experimental Setup}

\begin{figure*}[t]
    \centering
    \begin{subfigure}[b]{0.48\textwidth}
        \centering
        \includegraphics[width=\textwidth]{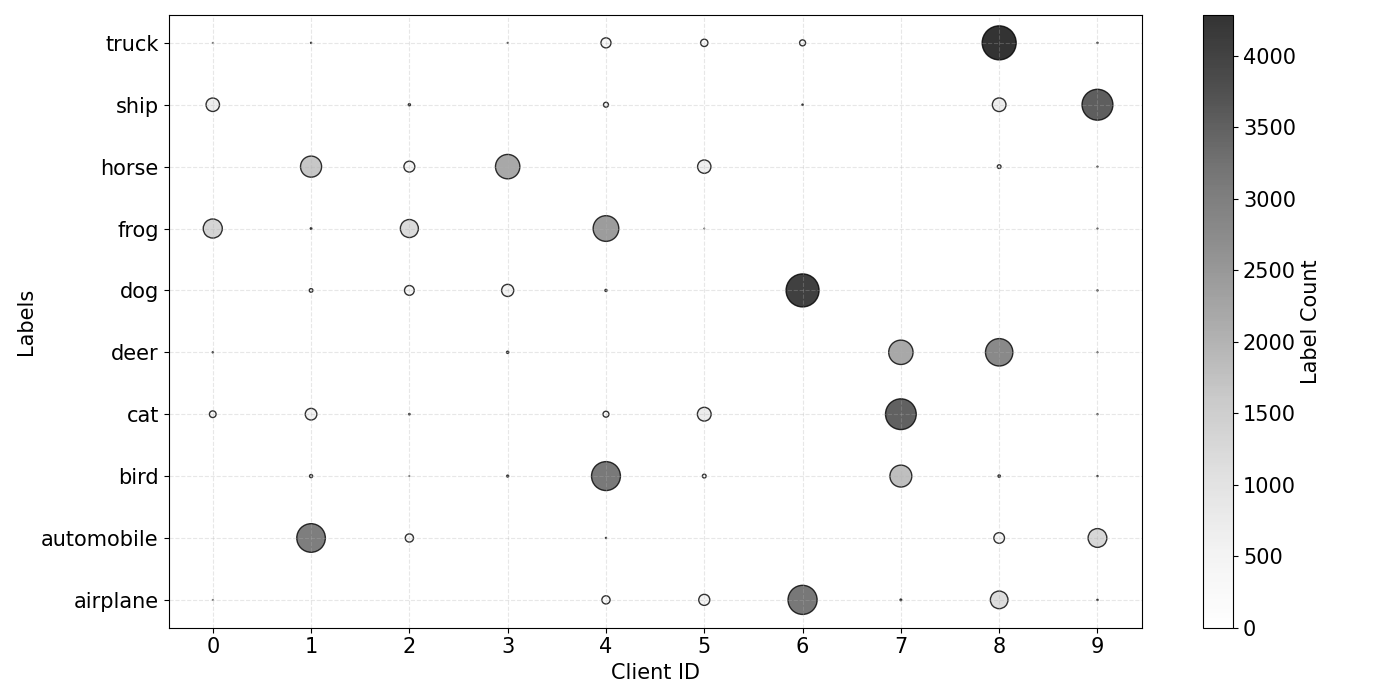}
        \caption{$a=0.1$}
        \label{fig:sub1}
    \end{subfigure}
    \hfill
    \begin{subfigure}[b]{0.48\textwidth}
        \centering
        \includegraphics[width=\textwidth]{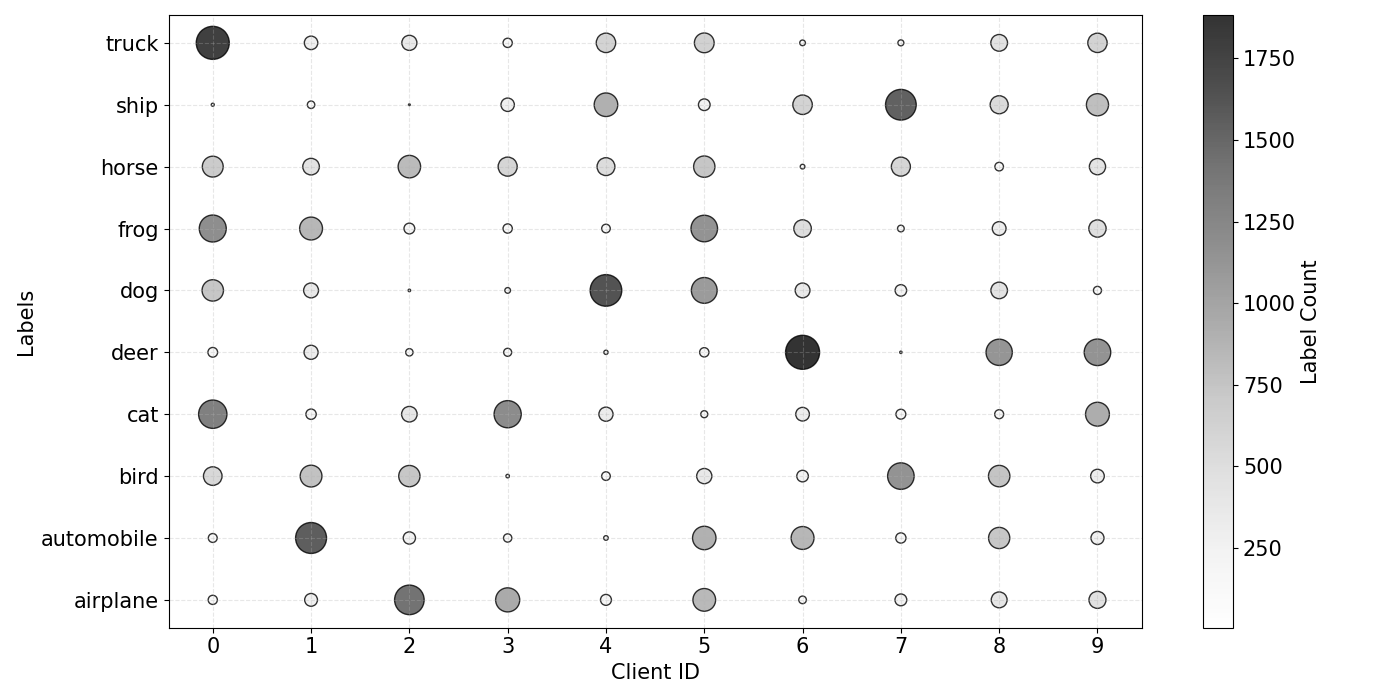}
        \caption{$a=1$}
        \label{fig:sub2}
    \end{subfigure}
    \caption{Illustration of data heterogeneity in the 10-client FL setup on CIFAR-10 using Dirichlet partitioning with $a=0.1$ (high skewness) and $a=1$ (mild skewness).}
    \label{fig:two_column_subfigures}
\end{figure*}

We validate the effectiveness of the proposed defense strategy by simulating an FL environment consisting of a central server and 10 clients collaboratively engaged in multi-label image classification training. Our evaluation leverages benchmark datasets commonly used in related literature \cite{lin2020ensemble, li2023experimental}, namely CIFAR-10\cite{krizhevsky2009learning}, FMNIST\cite{xiao2017fashion}, and MNIST\cite{yann2010mnist}. We emphasize non-IID data partitions to rigorously examine the robustness of our approach in comparison to the state-of-the-art defense methods, as described in Section~\ref{Resilient Aggregation Methods Against Byzantine Behavior}. 

In the examined non-IID setting, the training data of each dataset is partitioned across clients using a Dirichlet distribution, while the central server retains the test subset. Figure~\ref{fig:two_column_subfigures} visualizes the heterogeneous data distribution of the CIFAR-10 dataset across the 10 clients in our FL setup. Each circle denotes the number of samples for a given class assigned to a client, with larger circles indicating a higher label count in the client's local dataset. The figure captures two representative Dirichlet partitioning scenarios with concentration parameters $a=0.1$ (strongly skewed) and $a=1$ (moderately skewed), illustrating varying levels of heterogeneity in the data distribution.
MNIST and FMNIST datasets exhibit similar heterogeneous data distributions.

To support the implementation of FedGreed, the server-held dataset is evenly partitioned into two distinct subsets. The first subset is dedicated to evaluating the loss associated with each individual client's model update, as well as the loss arising from the intermediate aggregation of client updates, as specified in steps $(4)$ and $(5)$ of Algorithm~\ref{algGreedy} respectively. The second subset serves to evaluate the performance of the aggregated global model at the end of each FL communication round. Ensuring fairness and consistency, this evaluation set remains fixed across all compared defense strategies and is used to compute the centralized accuracy metric. As centralized accuracy, we measure the ratio of correct predictions to total predictions in the evaluation set at the end of each communication round.

We employ two convolutional neural networks (CNNs) for the model architecture, as defined by the Flower FL framework repository \cite{beutel2020flower}. One network is specifically designed for the CIFAR-10 dataset, while the other supports both the MNIST and FMNIST datasets. Within the scope of local client-level training, we investigate the performance of SGD and Adam optimizers, employing their default hyperparameter configurations. Our experimental results indicate that Adam achieves marginally superior accuracy compared to SGD, consistent with existing literature \cite{mills2021multi, reddi2020adaptive}, which highlights Adam’s enhanced convergence properties and overall efficacy in model optimization. Consequently, we adopt Adam as the preferred optimizer for the local client-side training.

Built upon the Flower framework, we conduct FL simulations over 50 communication rounds. Each simulation evaluates a specific defense mechanism under the presence of an adversarial attack. We enforce the examined defense scheme from the outset of training, while the adversarial attack is introduced at the 10th round and remains active until the end of the experiment. To assess the impact of Byzantine attacks, we consider three scenarios involving 3, 5, and 8 adversarial clients, respectively. Within each simulation, the attack type remains the same across all adversarial clients and corresponds to the threat models detailed in Section~\ref{Threat Models}. The defense mechanisms under evaluation include the proposed approach alongside baseline methods such as Mean, Trimmed Mean, Median, Krum, and Multi-Krum.

The majority of the assessed defense mechanisms necessitate parameter initialization. For Krum and Multi-Krum, prior knowledge of the maximum number of adversarial clients, denoted as $f$, is necessary; in our experiments, $f$ is set to match the actual number of adversarial clients. Additionally, Multi-Krum requires explicit determination of the cardinality of client updates included in aggregation; we establish this parameter to coincide with the population of benign participants. Regarding the Trimmed Mean method, the parameter $\beta$, governing the exclusion threshold for extreme values (both upper and lower bounds) in outlier removal, adopts its default setting of $\beta=0.2$.

This study explores the robustness and adaptive capacity of defensive mechanisms in preserving training stability and model integrity in adversarial FL environments aimed at degrading global model accuracy. Centralized accuracy serves as the primary evaluation metric across all defense-attack scenarios. In each experimental trial, adversarial attacks start at the 10th communication round and persist through the remainder of the simulation. We compute the mean centralized accuracy over these rounds to assess the effectiveness of each defense strategy, as detailed in the Experimental Results section. To support further exploration, the implementation used in our experiments is available in our public codebase\footnote{\url{https://github.com/Krith-man/FedGreed/}}.

\begin{table*}[t]
\caption{Centralized accuracy performance of defense mechanisms across three adversarial scenarios: \textbf{No Attack}, \textbf{Label Flipping}, and \textbf{Gaussian Noise} attacks, evaluated on three benchmark datasets in a heterogeneous 10-client FL setup. Results reflect the mean accuracy over three independent FL simulations, each executed with distinct random seeds for adversarial client sampling. Dirichlet concentration parameter is denoted by $(a)$, and the number of adversarial clients is indicated by $(M)$. Bold values denote the highest accuracy achieved for each dataset within a given attack scenario.}
\centering
\scriptsize
\setlength{\tabcolsep}{2pt}
\renewcommand{\arraystretch}{1.2}
\begin{tabular}{ll*{18}{c}}
\toprule
\multirow{4}{*}{\textbf{Defense Method}} & 
\multirow{4}{*}{\textbf{Attack Type}} & 
\multicolumn{18}{c}{\textbf{Dataset}} \\
\cmidrule(lr){3-20}
& & \multicolumn{6}{c}{\textbf{CIFAR-10}} & \multicolumn{6}{c}{\textbf{FMNIST}} & \multicolumn{6}{c}{\textbf{MNIST}} \\
\cmidrule(lr){3-8} \cmidrule(lr){9-14} \cmidrule(lr){15-20}
& & \multicolumn{2}{c}{$M{=}3$} & \multicolumn{2}{c}{$M{=}5$} & \multicolumn{2}{c}{$M{=}8$}
  & \multicolumn{2}{c}{$M{=}3$} & \multicolumn{2}{c}{$M{=}5$} & \multicolumn{2}{c}{$M{=}8$}
  & \multicolumn{2}{c}{$M{=}3$} & \multicolumn{2}{c}{$M{=}5$} & \multicolumn{2}{c}{$M{=}8$} \\
\cmidrule(lr){3-4} \cmidrule(lr){5-6} \cmidrule(lr){7-8}
\cmidrule(lr){9-10} \cmidrule(lr){11-12} \cmidrule(lr){13-14}
\cmidrule(lr){15-16} \cmidrule(lr){17-18} \cmidrule(lr){19-20}
& & $a{=}0.1$ & $a{=}1$ & $a{=}0.1$ & $a{=}1$ & $a{=}0.1$ & $a{=}1$
  & $a{=}0.1$ & $a{=}1$ & $a{=}0.1$ & $a{=}1$ & $a{=}0.1$ & $a{=}1$
  & $a{=}0.1$ & $a{=}1$ & $a{=}0.1$ & $a{=}1$ & $a{=}0.1$ & $a{=}1$ \\
\midrule

\multirow{2}{*}{FedGreed} & No Attack & 50.98 & 64.73 & 49.93 & 65.25 & 50.59 & 64.02 & 82.89 & 88.49 & 82.39 & 88.82 & 82.99 & 88.66 & 97.57 & 98.27 & 97.81 & 98.22 & 97.65 & 98.3 \\
& Label Flip & \textbf{43.52} & \textbf{63.16} & \textbf{37.55} & \textbf{61.56} & \textbf{17.42} & \textbf{52.47} & \textbf{77.78} & \textbf{87.31} & \textbf{72.35} & \textbf{86.6} & \textbf{52.32} & \textbf{82.66} & \textbf{95.77} & \textbf{97.92} & \textbf{94.19} & \textbf{97.91} & \textbf{67.88} & \textbf{96.81} \\
& Gaussian Noise & 10.15 & 63.35 & 10.07 & \textbf{61.62} & 10.21 & \textbf{32.75} & 19.39 & \textbf{87.72} & 11.68 & \textbf{87.17} & 11.25 & \textbf{83.71} & \textbf{95.47} & 98.08 & \textbf{93.59} & \textbf{97.93} & \textbf{14.11} & \textbf{96.89} \\
\midrule

\multirow{2}{*}{Multi-Krum} & No Attack & 50.89 & 64.26 & 50.23 & 64.3 & 51.15 & 64.59 & 81.99 & 88.92 & 82.94 & 88.45 & 83.01 & 88.7 & 97.73 & 98.31 & 97.77 & 50.43 & 97.64 & 98.19 \\
& Label Flip & 23.2 & 49.72 & 20.69 & 22.65 & 15.28 & 6.27 & 59.83 & 73.34 & 43.76 & 54.04 & 11.48 & 1.04 & 59.43 & 97.18 & 40.51 & 38.98 & 5.42 & 32.39 \\
& Gaussian Noise & \textbf{23.83} & \textbf{63.36} & 16.16 & 10.96 & 10.03 & 10.24 & \textbf{79.83} & 87.58 & \textbf{71.89} & 86.61 & 10 & 10 & 92.33 & \textbf{98.15} & 83.52 & 97.88 & 11.15 & 39.28 \\
\midrule

\multirow{2}{*}{Krum} & No Attack & 19.77 & 44.32 & 19.67 & 44.34 & 19.68 & 43.3 & 21.46 & 77.24 & 20.12 & 76.76 & 18.22 & 76.85 & 33.08 & 92.51 & 31.35 & 91.85 & 31.9 & 92.22 \\
& Label Flip & 16.39 & 34.44 & 19.1 & 9.67 & 14.06 & 6.6 & 23.52 & 75.26 & 12.9 & 57.23 & 11.47 & 1.03 & 30.11 & 92.31 & 23.67 & 94.36 & 8.02 & 31.39 \\
& Gaussian Noise & 15.95 & 52.51 & \textbf{18.87} & 46.73 & 10.3 & 10.3 & 31.43 & 75.82 & 33.74 & 75.72 & \textbf{12.85} & 10 & 33.61 & 92.41 & 24.34 & 94.24 & 14.08 & 38.31 \\
\midrule

\multirow{2}{*}{Median} & No Attack & 48.43 & 63.83 & 48.5 & 64.15 & 48.53 & 63.71 & 76.97 & 88.11 & 77.57 & 88.51 & 76.78 & 88.42 & 95.6 & 98.16 & 96.07 & 98.26 & 95.66 & 98 \\
& Label Flip & 32.47 & 55.37 & 24.22 & 33.59 & 14.12 & 8.62 & 73.01 & 82.26 & 51.82 & 45.17 & 5.54 & 1.51 & 85.15 & 94.64 & 47.26 & 42.07 & 4.35 & 0.79 \\
& Gaussian Noise & 12.53 & 14.26 & 10.14 & 10.48 & 10.37 & 10.46 & 64.83 & 81.64 & 16.6 & 12.13 & 10 & 10 & 88.63 & 97.02 & 10.05 & 9.8 & 9.79 & 9.8 \\
\midrule

\multirow{2}{*}{Trimmed Mean} & No Attack & 48.85 & 64.65 & 49.43 & 63.63 & 49.87 & 64.31 & 80.67 & 88.5 & 79.79 & 88.36 & 79.92 & 88.63 & 96.79 & 98.19 & 96.51 & 98.26 & 96.6 & 98.27 \\
& Label Flip & 34.5 & 56.21 & 24.85 & 33.66 & 14.67 & 8.71 & 75.47 & 82.28 & 53.73 & 46.77 & 6.2 & 1.27 & 86.85 & 94.07 & 46.96 & 37.15 & 3.6 & 1.03 \\
& Gaussian Noise & 11.4 & 12.77 & 9.87 & 10.65 & 10.16 & 10.84 & 43.7 & 48.2 & 9.93 & 10 & 10 & 10 & 15.52 & 36.71 & 10.07 & 10.21 & 9.79 & 10.46 \\
\midrule

\multirow{2}{*}{Mean} & No Attack & 49.54 & 65.26 & 50.11 & 64.85 & 51.22 & 64.83 & 82.29 & 88.8 & 82.93 & 88.69 & 83.02 & 88.88 & 97.41 & 98.22 & 97.53 & 98.34 & 97.57 & 98.21 \\
& Label Flip & 34.83 & 53.84 & 25.76 & 30.87 & 13.28 & 9.35 & 63.58 & 81.48 & 38.48 & 32.7 & 7.41 & 1.84 & 83.75 & 96.45 & 63.26 & 66.84 & 18.39 & 1.07 \\
& Gaussian Noise & 10.09 & 10.98 & 10.64 & 10.79 & \textbf{10.81} & 10.91 & 17.23 & 28.9 & 13.95 & 10 & 10 & 10 & 18.04 & 10.41 & 9.39 & 10.11 & 9.24 & 10.92 \\
\bottomrule
\end{tabular}
\label{tab:average_rankings}
\end{table*}
\subsection{Experimental Results}

Table~\ref{tab:average_rankings} reports the mean centralized accuracy of all evaluated defense mechanisms across 3 datasets and under 3 threat model scenarios: no attack, label flipping, and Gaussian noise injection.

Each adversarial scenario is further examined under two levels of data heterogeneity, introduced via Dirichlet-distributed data splits: one with mild skewness ($a=1$) and another with high skewness ($a=0.1$) across participating clients. Additionally, we analyze the impact of varying the number of malicious clients ($M$), with adversarial behavior activated at the 10th communication round of each FL simulation.

In a heterogeneous FL environment, it is essential to evaluate the centralized accuracy of each simulation across different adversarial scenarios to ensure a holistic assessment \cite{lin2020ensemble}. Towards this direction, the reported results represent the mean centralized accuracy across three independent FL simulations, each conducted using distinct random seeds for adversarial client selection.

The experimental results presented in Table~\ref{tab:average_rankings} provide a systematic comparison between FedGreed and existing aggregation strategies. First, our method exhibits superior performance over the Mean, Trimmed Mean, and Median baselines under varying malicious client ratios, heterogeneity levels, and evaluated datasets. The proposed defense mechanism addresses a fundamental limitation of conventional approaches: their failure to reliably distinguish between benign and adversarial updates. Unlike these methods, which either naively aggregate all client updates or apply simplistic statistical trimming, our technique ensures a more robust and discriminative aggregation process by employing a greedy selection of model updates for aggregation in the server side.

In contrast, the Krum and Multi-Krum defense mechanisms employ a selective aggregation strategy predicated on pairwise Euclidean distance analysis. These methods rely on the fundamental assumption that model updates from honest clients demonstrate greater mutual similarity, evidenced by smaller Euclidean distance, relative to those from adversarial clients. Although this geometric property mostly results in improved mean centralized accuracies compared to Mean, Trimmed Mean and Median defenses, FedGreed surpasses Krum and Multi-Krum in most scenarios. Unlike our approach, Krum selects a single client update, specifically, the one with the minimal cumulative distance to the others within the selected subset, as the new global model, rather than aggregating updates from all its closed distances clients. Consequently, Krum does not fully exploit the informative contributions of all honest participants. Furthermore, under highly skewed FL environments, the assumption of close Euclidean distances among honest clients deteriorates, causing Multi-Krum to distinguish between honest and malicious users inadequately. An additional advantage of FedGreed is that, unlike Krum and Multi-Krum, which require a priori information regarding the maximum number of malicious clients to compute their distance scores, our approach operates without necessitating such information.

\begin{figure}[ht]
    \centering
    \begin{subfigure}{\textwidth}
        \includegraphics[width=0.48\linewidth]{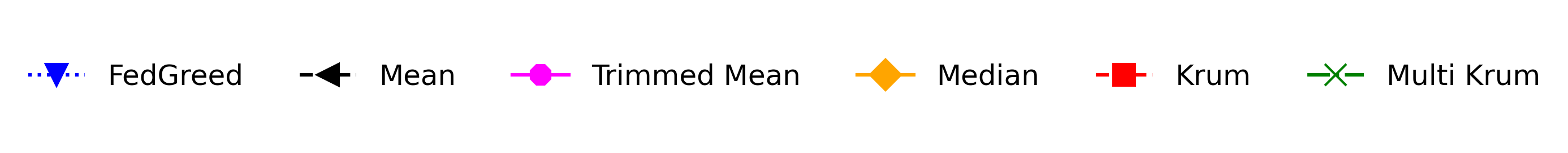}
    \end{subfigure}
    \begin{subfigure}{\textwidth}
        \includegraphics[width=0.48\linewidth]{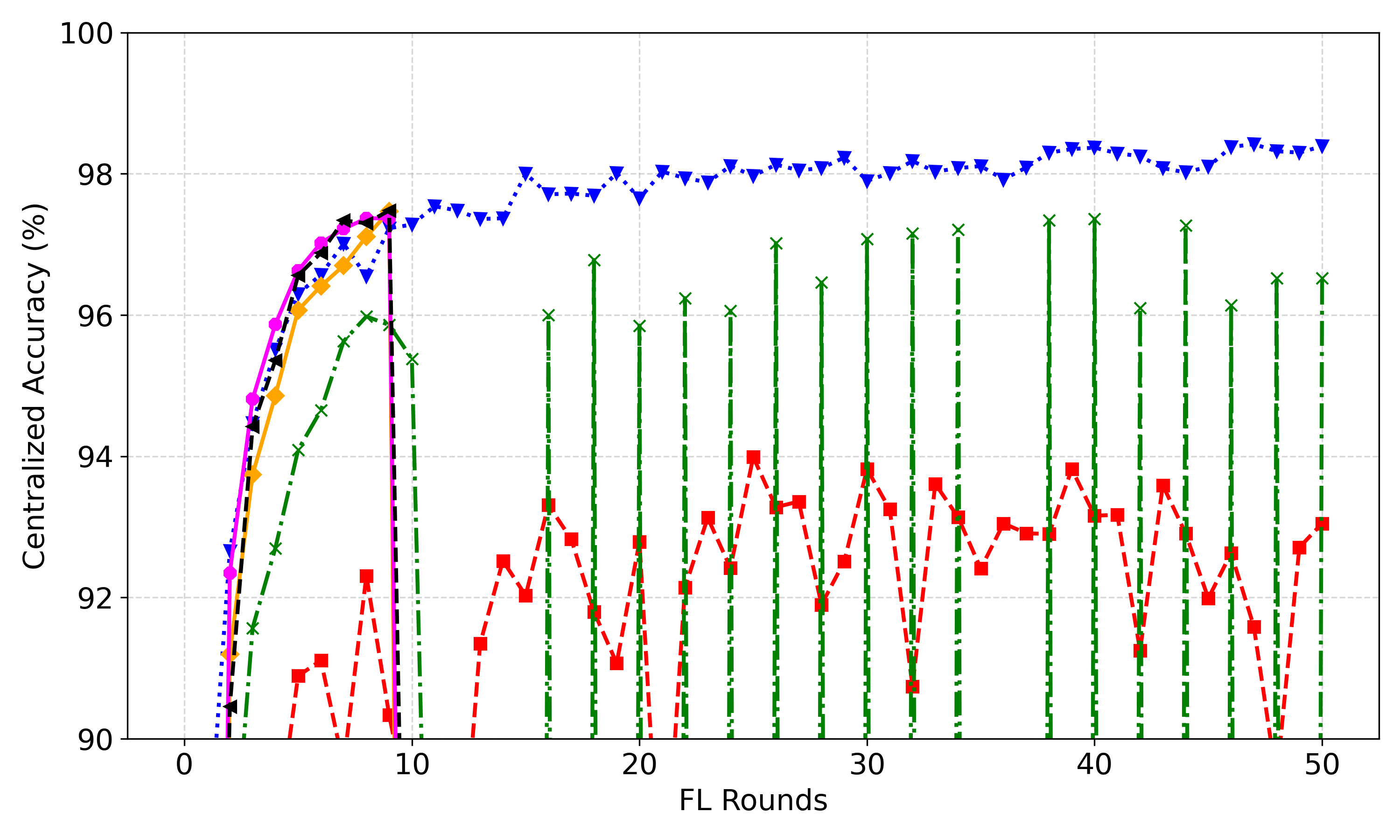}
    \end{subfigure}

    \caption{Centralized accuracy over 50 FL rounds on MNIST, comparing all 6 defense methods against Label Flipping attack in a mild heterogeneous 10-client FL setup \((a=1)\) with 50\% malicious participants.}
    \label{fig:2}
\end{figure}

Figure~\ref{fig:2} presents the performance of all 6 evaluated defense methods in terms of centralized accuracy in the MNIST dataset under a Label Flipping attack within an FL environment comprising 10 clients, 5 of which are randomly designated as malicious. The entire FL experiment spans 50 communication rounds. While the defense mechanisms are applied from the beginning of each simulation, the Label Flipping attack is introduced at the 10th communication round. The results support prior findings: the Median, Mean, and Trimmed Mean defenses experience a substantial decline in accuracy following the onset of the attack. Although Krum demonstrates resilience against the adversarial behavior, its performance remains limited and does not match the accuracy levels achieved by FedGreed. Multi-Krum yields results closer to those of our proposed method; however, its inconsistency in correctly identifying honest clients results in abrupt drops in centralized accuracy throughout the FL rounds. 

Focusing on the adversarial attacks applied to FedGreed across the two heterogeneous environments considered, we observe that the proposed method exhibits vulnerability in a limited number of highly skewed cases, most notably under the Gaussian Noise attack. By design, FedGreed ensures that, at each iteration, the selected update yields progress at least comparable to that of the client with the most favorable value of $f_{\mathrm{S}}$. However, in scenarios with extreme data skewness, the progress made by the best-performing individual client with respect to 
$f_{\mathrm{S}}$ may be minimal. In such cases, the gradient direction of a single client can significantly deviate from the average gradient direction of all honest clients—the desired update direction. Consequently, this individual gradient may resemble random noise. When an adversarial client produces loss values similar to those of highly skewed honest participants, it becomes possible for a malicious update to be selected as the one minimizing the aggregated loss. So even though we guarantee bounded error, this bound depends explicitly on the gradient noise of individual clients, and in high skewness, this bound is so large that it manifests in non-useful accuracies at testing. We acknowledge this observation as a direction for future work, where we aim to address and mitigate this rare vulnerability in our results.

\section{Conclusions}
\label{conclusions}

In this work, we introduced FedGreed, a greedy client selection algorithm tailored for highly adversarial and heterogeneous FL environments. FedGreed ranks clients' local model updates based on their loss values evaluated by the server-side trusted dataset and adaptively selects, in a greedy manner, a subset of clients whose models achieve the lowest evaluation loss. This approach does not require any assumption on the number of malicious clients and remains effective even in highly adversarial settings. 

Experimental results on benchmark datasets, including MNIST, FMNIST, and CIFAR-10, indicate that our method consistently surpasses both standard and robust aggregation baselines, such as Mean, Trimmed Mean, Median, Krum, and Multi-Krum, across the majority of adversarial scenarios examined. This enhanced performance in centralized accuracy persists even under adversarial conditions with up to 80\% malicious clients. Notably, our defense remains resilient across varying levels of data heterogeneity and attack types, including label flipping and Gaussian noise injection. As future work, we plan to extend this framework to scenarios with partial client participation, more complex attack patterns, and integration into real-world federated systems with dynamic trust management.
\section*{Acknowledgment}
\label{acknowledgment}

This work has received funding from the European Union’s Horizon Europe Research and Innovation Programme under grant agreement No. 101168560 (CoEvolution). This paper reflects only the authors’ view and the Commission is not responsible for any use that may be made of the information it contains. The work of D. Jakovetic was supported by the Ministry of Science,
Technological Development and Innovation of the Republic of Serbia
(Grants No. 451-03-137/2025-03/200125 \& 451-03-136/2025-03/200125),
and by the Science Fund of the Republic of Serbia, Grant no. 7359,
project LASCADO.

\bibliographystyle{IEEEtran}
\bibliography{references} 

\end{document}